\documentclass[sigconf,natbib=true]{llncs}

\usepackage[T1]{fontenc}
\usepackage{graphicx}
\usepackage{amsfonts}
\usepackage{amsmath}
\usepackage[hidelinks]{hyperref}
\usepackage{booktabs}
\usepackage{subcaption}
\usepackage[disable]{todonotes}
\usepackage{multirow}
\usepackage{graphicx}
\usepackage{fixfoot}
\usepackage{orcidlink}
\usepackage[normalem]{ulem}
\usepackage{float}
\usepackage{arydshln}
\usepackage{wrapfig}

\begin{document}
\newcommand{\experimentRepo}{\url{https://github.com/wendli01/robust_link}}

\newcommand{\new}[1]{\textcolor{olive}{#1}}
\renewcommand{\new}[1]{#1}

\definecolor{greenish}{HTML}{82B366}
\definecolor{blueish}{HTML}{6C8EBF}
\definecolor{orangeish}{HTML}{D79B00}
\definecolor{redish}{HTML}{B85450}
\definecolor{bluegray}{HTML}{647687}

\DeclareFixedFootnote{\osf}{\url{https://osf.io/8d2v4/}; March 30, 2025}
\DeclareFixedFootnote{\rgcn}{The large test time difference is due to the optimized RGCN implementation in \texttt{DGL}}

\title{Robust Generalizable Heterogeneous Legal Link Prediction}
\titlerunning{Robust Legal Link Prediction}

\author{Lorenz Wendlinger\orcidlink{0000-0001-9459-6244}\textsuperscript{1} \and
Simon Alexander Nonn\orcidlink{0009-0007-7173-5302}\textsuperscript{1}\and
Abdullah Al Zubaer\orcidlink{0009-0001-0842-7434}\textsuperscript{1}\and
Michael Granitzer\orcidlink{0000-0003-3566-5507}\textsuperscript{1,2}}

\authorrunning{L. Wendlinger et al.}

\institute{\textsuperscript{1} Universit\"at Passau, Passau, Germany \\
\textsuperscript{2} Interdisciplinary Transformation University Austria, Linz, Austria
	\email{\{lorenz.wendlinger,simon.nonn,abdullahal.zubaer,michael.granitze\}@uni-passau.de}
}

\maketitle 

\begin{abstract}
Recent work has applied link prediction to large heterogeneous legal citation networks \new{with rich meta-features}.
We find that this approach can be improved by including edge dropout and feature concatenation for the learning of more robust representations, which reduces error rates by up to 45\%.
We also propose an approach based on multilingual node features with an improved asymmetric decoder for compatibility, which allows us to generalize and extend the prediction to more, geographically and linguistically disjoint, data from New Zealand.
Our adaptations also improve inductive transferability between these disjoint legal systems.

\end{abstract}
\section{Introduction}


Link prediction is an important application in legal tech, as it goes beyond mere reference extraction and information retrieval by directly predicting missing links using semantic and topological features.
Recent work by Wendlinger et al. \cite{wendlinger2025missing} has shown that it can be applied effectively to large heterogeneous citation graphs by performing joint prediction on a meta-feature enriched graph with a convolution mechanism adapted to their scale-free nature.
The meta-feature enrichment process injects additional information, which can alleviate the effects of intrinsically or extrinsically missing links.
However, they also make the model susceptible to erroneous or spurious meta-information, which can arise from data quality issues or application limitations.
\new{Their enhanced relational graph convolution with a relational self-loop is show to effectively handle this additional topological information.}

Furthermore, the preferential attachment observed in citation networks together with the tendency to over-parameterize in heterogeneous models can promote over-fitting.
To alleviate these problems, and promote the learning of robust node representations, we propose incorporating edge dropout \cite{rong2020dropedgedeepgraphconvolutional}.
This simulates the impact of missing edges, which is common in this domain, during training and prevents problematic adaptation to spurious links.
We further incorporate the representation concatenation scheme from \cite{lv2021we}, along with a compatible asymmetric decoder, which provides a directly accessible multi-hop view of topology that can complement the residual connections employed in \cite{wendlinger2025missing}.
We show the capability of our adaptations against an extended model cohort, including GNNs expressly designed for inductive learning (GraphSAGE) and heterogeneous graphs (simple HGN, HGT).

While the evaluation setting of \cite{wendlinger2025missing} is robust and semi-inductive by nature, fully inductive and transfer learning are not explored, and only one base graph from on one legal system is considered.
This limits their findings to the civil law focused German legal system, which may not translate to the patterns of the more common case law.
\todo{remove remarks on evaluation if they don't fit}
We extend this by investigating performance on a geographically, legally, and linguistically diverse legal citation graph from New Zealand \new{that is similar in flavour}, LiO338k \cite{milz2023law}.
We further conduct a study of 
fully inductive transfer between the legal systems of Germany and New Zealand.
We also include more relevant models from related work, that have been specifically designed for similar domains, such as Graph SAGE \cite{hamilton2017inductive} for inductive learning, HGE\,\cite{hu2020heterogeneous} for heterogeneous attention, and the simple HGN explicitly proposed as an alternative\,\cite{lv2021we}.


\section{Robust Generalizable Legal Link Prediction}
\label{sec:abres-gcn}

\begin{figure}[h]
  \begin{center}
    \includegraphics[width=\linewidth,height=.5\textheight,keepaspectratio]{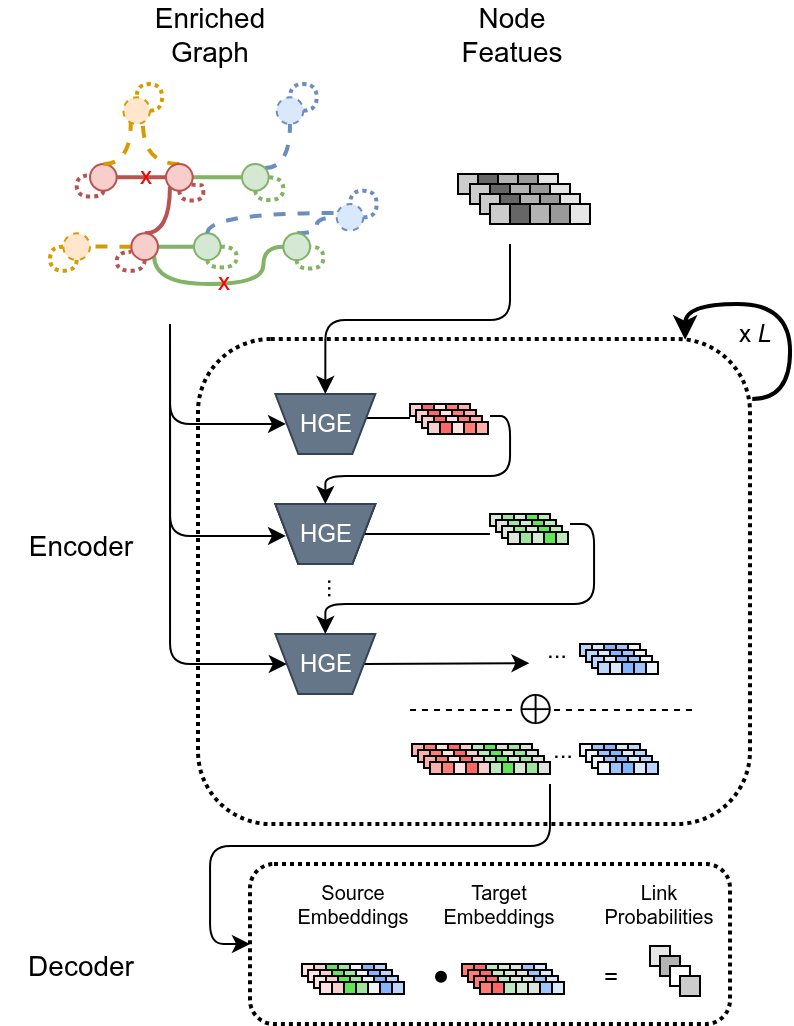}
  \end{center}
  \caption{R-HGE for a graph with some dropped edges (\textcolor{red}{x}), feature concatenation and asymmetric interleave decoder.\vspace{-0ex}}
  \label{fig:rhge_flowchart}
\end{figure}

Our Robust Heterogeneous Graph Enrichment Model (\textbf{R-HGE}) improves upon previous work \cite{wendlinger2025missing} in terms of robustness and prediction quality, c.f. \autoref{fig:rhge_flowchart}.

\textbf{Edge Dropout} 
We employ edge dropout from \cite{rong2019dropedge} as an augmentation strategy, in addition to feature dropout, with a probability of 50\% to improve robustness as well as outright prediction quality through regularization.
This strategy applies well to legal link prediction as it alleviates the reliance on specific edges that can be spurious or missing due to data quality issues.
We apply edge dropout to all target relations and their reverse to replicate the impact of missing edges, but not missing meta-information represented through respective relations.
This interacts well with the enrichment procedure of HGE, by not introducing noise in removing meta-information.

\textbf{Feature Concatenation}
Previous work \cite{bresson2017residual,xu2018representation} describes the utility of skip connections for residual computation, as they can improve the variability in receptive field size \todo{add citation)}.
Feature concatenation \cite{xu2018representation,lv2021we} further helps reduce over-fitting by directly exposing all layer representations, and therefore topological levels, to the decoder.

\textbf{Decoding}
While \cite{wendlinger2025missing} use the asymmetric inner product decoder from \cite{yu2014link} to generate link likelihoods from node representations, we rearrange this to extract source and target representations by interleaving instead of block-wise.
This ensures compatibility with semantic dimensions in node features, such as in our concatenated representations.


\section{Datasets}
\label{sec:datasets}

We explore link prediction performance on the two legal citation graphs OLD201k \cite{kdir21} and LiO338k \cite{milz2023law}, which are similar in flavour but extracted from fully disjoint databases.
\new{They are unique in both their heterogeneity as well as the availability of full textual content for all nodes.
Both data sets contain case publication dates, allowing for semi-inductive temporal splitting and therefore more realistic evaluation compared to random splitting.
This, we feel, is a better representation of real-life citation prediction performance than the fully transductive settings of related work \cite{lv2021we,liu2024msgnn}}.
For graph composition statistics, see \autoref{tab:dataset_stats}.
Both graphs are structurally similar; while LiO338k is larger with respect to the included cases and variety of courts, it features fewer distinct law texts.
Consequently, where norm references outnumber case references 10 to 1 in OLD201k, they are almost balanced for LiO338k, reflecting the orientation of the two legal systems towards civil law and case law respectively.

\begin{table*}
    \centering
    \begin{tabular}{c|c|r|r|c}
    \toprule
        Type & Name & OLD201k&LiO338k& meta-information  \\
        \midrule
        \multirow[c]{4}{*}{Node} & Case & 201\,823 & 338\,360 & type, date\\
        & Law & 50\,814 & 10\,402 &law book code, law book title, section\\
        & \multirow[c]{2}{*}{Court} & \multirow[c]{2}{*}{124} & \multirow[c]{2}{*}{1\,119} & name, type, slug, city, description\\
        &&&&state, jurisdiction, level of appeal\\\midrule
        \multirow[c]{3}{*}{Edge} & \textbf{C}ase-\textbf{C}ase* & 90\,189& 236\,458& -\\
        & \textbf{C}ase-\textbf{L}aw* & 971\,625& 244\,159&-\\
        & \textbf{L}aw-\textbf{L}aw & - & 67\,857&-\\
        & \textbf{L}aw-\textbf{C}ase & - & 62&-\\
        \bottomrule
        
    \end{tabular}
    \caption{Dataset statistics for OLD201k and LiO338k. *used as prediction targets}
    \label{tab:dataset_stats}
\end{table*}

\begin{figure}
    \begin{subfigure}[t]{.5\linewidth}
        \centering
        \includegraphics[width=\linewidth]{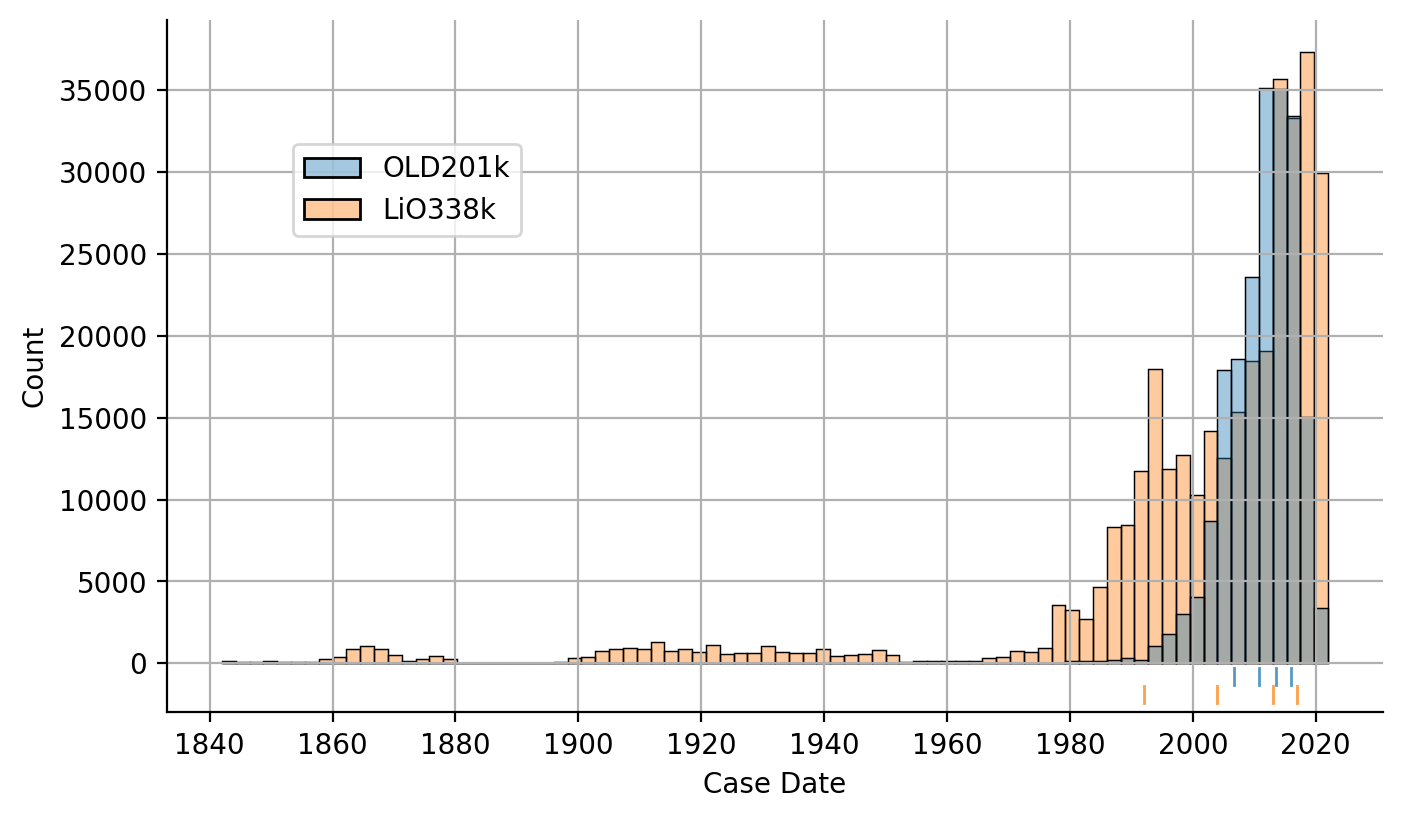}
        \caption{Case date with cutoffs for date splitting indicated for \textcolor{orangeish}{OLD201k} and \textcolor{blueish}{OLD36k}}
        \label{fig:date_dist}
    \end{subfigure}
    \hfill
    \begin{subfigure}[t]{.5\linewidth}
        \centering
        \includegraphics[width=\linewidth]{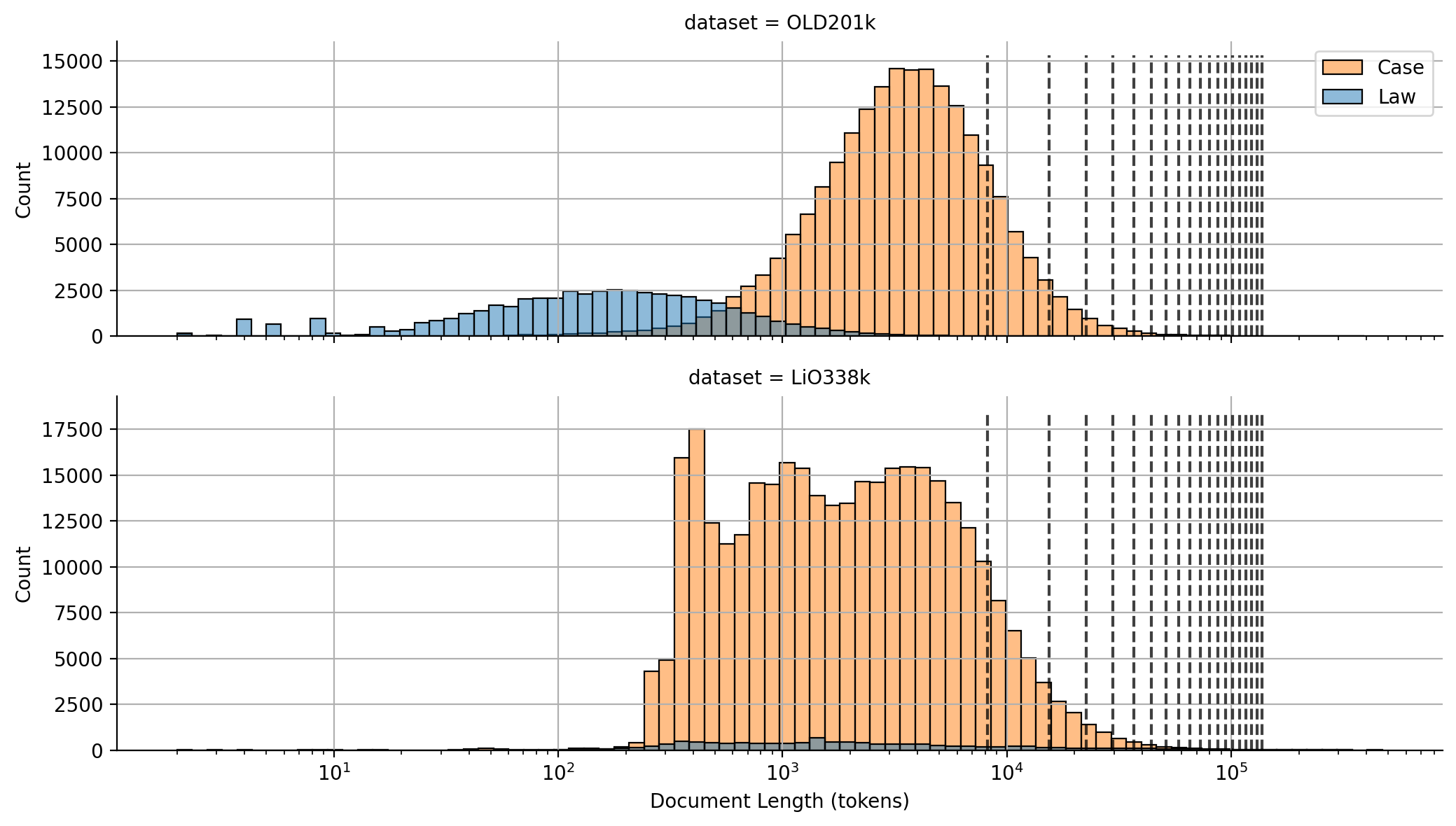}
        \caption{Document Length with embedding context window \dashuline{marked}}
        \label{fig:doc_length}
    \end{subfigure}
    \caption{OLD201k and LiO338k Dataset Distributions.}
    \label{fig:old_dist}
\end{figure}

Both graphs have a long historic tail (c.f. \autoref{fig:old_dist}), though only the cases from LiO338k go back almost 2 centuries.
LiO338k further contains shorter case documents, with a peak around 400 tokens, while OLD201 case lengths are approximately normally distributed.
Conversely, New Zealand laws in LiO338k are much more verbose and less fragmented than those in OLD201k.

\section{Experiments}

Here we present link prediction results for previous methods and our R-HGE on OLD201k and LiO338k in \autoref{subsec:main_study} as well as performing an ablation study to bridge the gap between HGE and R-HGE in \autoref{subsec:ablation_study}.
Finally, \autoref{subsec:robustness} shows  model robustness to topological attack and transfer between legal systems.
We keep the semi-inductive time-based graph splitting with cumulative testing and training from \cite{wendlinger2025missing} as well as their joint prediction scenario by default.

Like \cite{wendlinger2025missing}, we use latent sentence embeddings as node features, \texttt{jina-v2-base-de} embeddings \cite{günther2024jinaembeddings28192token}
 for OLD201K and \texttt{jina-v2-base-en} embeddings \cite{günther2024jinaembeddings28192token}
 for LiO338k.
For transfer, the multilingual \texttt{jina-v3} embeddings \cite{sturua2409jina}
 are used
 .



All models are trained by optimizing the cross-entropy reconstruction loss via adaptive momentum\,\cite{kingma2014adam} gradient descent 
for 200 epochs with a learning rate of $10^{-4}$.
We use 3 layers of size 256 for all models and 2 attention heads for GAT, HGT and simple HGN.
We implement all models in \texttt{DGL 2.4.0}
using \texttt{pytorch 2.6.0} 
 compiled with \texttt{CUDA 12.4} and run experiments on an Nvidia A100 GPU.
All code \new{with re-usable estimators}, experiment setups, final and intermediate results are available at \experimentRepo.


\subsection{Link Prediction}
\label{subsec:main_study}

We extend legal citation prediction to the case law legal system of New Zealand in \autoref{tab:main_study}.
In addition to the SGD, GCN\,\cite{kipf2016semi}, RGCN\,\cite{schlichtkrull2018modeling} and HGE\,\cite{wendlinger2025missing} models tested in \cite{wendlinger2025missing}, we also include strong baseline GNNs with Graph SAGE \cite{hamilton2017inductive} for its strength in inductive learning.
We also compare to the HGT\,\cite{hu2020heterogeneous} as its factorization of relations should fit the data well and the simple HGN of \cite{lv2021we}, which the authors show to out-perform many more complex GNNs.
We find LiO338k to be a significantly easier task, due to the inclusion of more of the generally more challenging case references, \new{as well as the richer topology afforded by the Law-Law and Law-Case links}.

\begin{table*}
\centering
    \begin{tabular}{l|ll|ll|ll|ll}
    \toprule
    Data& \multicolumn{4}{c|}{OLD201k} &\multicolumn{4}{c}{LiO338k}\\
    & &AUC- & \multicolumn{2}{c|}{time(s)}& &AUC- & \multicolumn{2}{c}{time(s)}\\
     Model & AP & ROC &  test & train & AP & ROC &  test & train \\
    \midrule
        SGD & 72.1$_{\pm0.92}$ & 72.6$_{\pm1.2}$ & 372$_{\pm330}$ & 211$_{\pm238}$ & 84.2$_{\pm2.4}$ & 87.7$_{\pm1.8}$ & 174$_{\pm160}$ & 155$_{\pm114}$ \\
        \hdashline
        s-HGN & 80.3$_{\pm2.3}$ & 79.1$_{\pm2.5}$ & 2.8$_{\pm1.3}$ & 171$_{\pm127}$ & 84.8$_{\pm8.1}$ & 83.7$_{\pm9.2}$ & 2.4$_{\pm1.3}$ & 95$_{\pm63}$ \\
        GCN & 80.8$_{\pm1.3}$ & 82.1$_{\pm1.3}$ & 2.9$_{\pm1.3}$ & 140$_{\pm103}$ & 91.4$_{\pm2.9}$ & 89.4$_{\pm4.9}$ & 2.2$_{\pm1.2}$ & 60$_{\pm45}$ \\
        RGCN & 85.6$_{\pm1.8}$ & 83.1$_{\pm2.0}$ & 1.2$_{\pm0.95}$ & 140$_{\pm101}$ & 90.7$_{\pm1.3}$ & 87.8$_{\pm2.7}$ & 2.0$_{\pm1.2}$ & 61$_{\pm51}$ \\
        HGT & 86.4$_{\pm1.2}$ & 84.6$_{\pm2.0}$ & 3.7$_{\pm1.8}$ & 313$_{\pm168}$ & 92.5$_{\pm1.3}$ & 90.9$_{\pm1.3}$ & 2.5$_{\pm1.5}$ & 244$_{\pm105}$ \\
        SAGE & 83.7$_{\pm2.3}$ & 84.8$_{\pm2.1}$ & 2.8$_{\pm1.2}$ & 134$_{\pm100}$ & \textit{95.5}$_{\pm1.7}$ & \textit{96.1}$_{\pm1.3}$ & 2.0$_{\pm1.2}$ & 63$_{\pm44}$ \\
        HGE & \textit{88.0}$_{\pm1.3}$ & \textit{85.8}$_{\pm1.6}$ & 4.3$_{\pm1.3}$ & 172$_{\pm100}$ & 93.6$_{\pm0.36}$ & 91.5$_{\pm0.81}$ & 2.6$_{\pm1.3}$ & 156$_{\pm87}$ \\
        \hdashline
        R-HGE & \textbf{91.0}$_{\pm1.4}$ & \textbf{90.3}$_{\pm1.7}$ & 4.1$_{\pm1.2}$ & 238$_{\pm136}$ & \textbf{97.5}$_{\pm0.48}$ & \textbf{97.4}$_{\pm0.61}$ & 2.7$_{\pm1.4}$ & 183$_{\pm90}$ \\
        \bottomrule
    \end{tabular}
    \caption{Link prediction results in 5 folds with date-based splitting and evaluation on 5 90\% test splits. Results $\pm\sigma$ are reported in \%  with macro-averaging over the edge types. GCN and GraphSAGE operate on the fully homogeneous bi-directed graph, SGD only on node embeddings.
    }
    \label{tab:main_study}
\end{table*}

Our robust R-HGE out-performs HGE by a significant margin in both metrics on LiO338k, reducing error rates by at least $2.5 \times$, and approaching perfect prediction results.
Prediction on LiO338k benefits greatly from edge dropout or neighborhood sub-sampling, as evidenced by the excellent GraphSAGE results.
Homogeneous models are competitive, with the simple GCN \cite{kipf2016semi} outperforming the heterogeneous RGCN \cite{schlichtkrull2018modeling}, as well as the other homogeneous methods.

This is in contrast to OLD201k, where heterogeneous methods surpass other approaches and the baseline, as well as the homogeneous GraphSAGE.
In this setting, too, R-HGE improves over HGE \cite{wendlinger2025missing}, most notably in AUC-ROC, hinting at much higher ranking consistency.
While the heterogeneous graph transformer consistently surpasses the heterogeneous, but attention less, RGCN, it still lags behind the more specialized inductive and enrichment based models.
While simple HGN \cite{lv2021we} is expressly designed to replace more complex heterogeneous approaches, with edge type embeddings providing context, it does not do well in our semi-inductive setting.

\new{R-HGE adds a slight computational burden over HGE due to the large representations, but is still more efficient than HGT.}

\subsection{Ablation Studies}
\label{subsec:ablation_study}

To bridge the gap between R-HGE and HGE, we investigate the precise impact of our improvements in \autoref{tab:ablation_study}, and find it to be very consistent between datasets.
While the asymmetric interleave decoding alone has negligible impact, as it is only a compatibility improvement, feature concatenation is beneficial, but cannot fully substitute residual connections.
The largest difference is introduced with the improved robustness afforded through training with edge dropout.

\begin{table}
\centering
    \begin{tabular}{l|ll|ll}
    \toprule
    Data& \multicolumn{2}{c|}{OLD201k} &\multicolumn{2}{c}{LiO338k}\\
     & AP & AUC-ROC  & AP & AUC-ROC \\
    \midrule
        HGE & 88.0$_{\pm1.3}$ & 85.8$_{\pm1.6}$ & 93.5$_{\pm0.31}$ & 91.4$_{\pm0.74}$ \\
        + interleave & 87.8$_{\pm1.4}$ & 85.4$_{\pm1.7}$ & 93.2$_{\pm0.72}$ & 90.9$_{\pm1.2}$ \\
        + cat all & 88.5$_{\pm1.5}$ & 86.6$_{\pm1.8}$ & 93.9$_{\pm0.56}$ & 92.1$_{\pm0.9}$ \\
        + edge drop & 90.7$_{\pm1.3}$ & 90.0$_{\pm1.6}$ & 96.8$_{\pm0.23}$ & 96.5$_{\pm0.33}$ \\\hdashline
        - residual & 90.8$_{\pm1.4}$ & 90.2$_{\pm1.7}$ & 97.1$_{\pm0.48}$ & 96.9$_{\pm0.59}$ \\
        - interleave & 90.9$_{\pm1.2}$ & 90.2$_{\pm1.4}$ & 97.0$_{\pm0.33}$ & 96.7$_{\pm0.45}$ \\
        R-HGE & 91.1$_{\pm1.4}$ & 90.4$_{\pm1.6}$ & 97.5$_{\pm0.47}$ & 97.4$_{\pm0.59}$ \\
        \bottomrule
    \end{tabular}
    \caption{R-HGE ablation study for OLD201k and LiO338k link prediction in 5 folds with date-based splitting and evaluation on 5 90\% test splits.}
    \label{tab:ablation_study}
\end{table}

\subsection{Transferability}
\label{subsec:robustness}

\begin{figure}[ht]
    \centering
    \includegraphics[width=\linewidth]{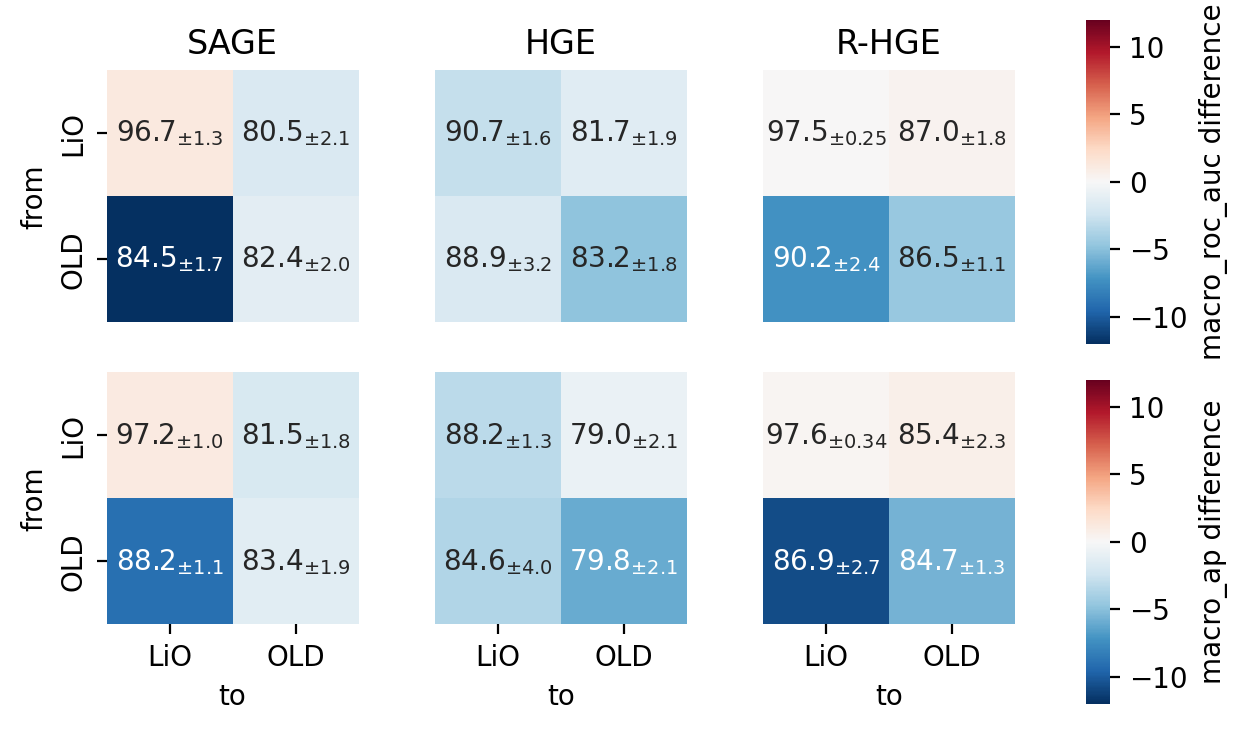}
    \caption{Fully inductive transfer within and between datasets, colour indicates deviation from the semi-inductive score for on-diagonal elements and from the within-dataset score otherwise.}
    \label{fig:transfer}
\end{figure}
\todo{Add GCN or GraphSAGE results? Remove one of the metrics as they correlate anyway?}

We test transferability between datasets, and their respective legal domains, with splitting over time, in a fully inductive scenario.
For HGE and R-HGE, meta-feature graph enrichment is deactivated, as it cannot handle the different meta-features between OLD201k and LiO338k and we use the multilingual \texttt{jina-V3} embeddings as node features
.

Out of the three models, the GraphSAGE model developed specifically for inductive learning suffers the least going from semi-inductive to fully inductive though R-GCN still out-performs it, c.f. \autoref{fig:transfer} as compared to \autoref{tab:main_study}. 
HGE does not adapt well to this scenario, with GraphSAGE and HGE performing similarly on OLD201k despite HGE's better semi-inductive results (c.f. \autoref{tab:main_study}), while GraphSAGE even improves slightly on LiO338k.
Despite this, R-HGE cannot effectively transfer from OLD201k to LiO338k, though it still offers the best results.
GraphSAGE behaves similarly while HGE is less affected.
Transfer from LiO338k to OLD201k is more positive throughout, with R-HGE improving slightly and HGE and GraphSAGE scores degrading slightly.

This shows that, while R-HGE can generalize well, even in inductive settings and from a densely sampled legal domain to a sparser one, it breaks down when extrapolating from fewer training data, while still outperforming other models.

\section{Conclusion}

We improve upon the state-of-the-art in legal link prediction using edge dropout and feature concatenation with adapted decoding for robustness and generally enhanced prediction.
We also extend the evaluation to a large disjoint citation network from a different legal system as well as fully inductive and transfer studies, where our robust model generalizes well.
This confirms that our adaptations pair well with the heterogeneous graph enrichment developed for legal citation prediction and can make it more robust and adaptable to dissimilar legal data.
While our study still has some limitations regarding data quality and availability, we consider this an important step towards a more robust and general link prediction system.


\subsubsection*{Acknowledgements}
The paper has been partially funded by COMET K1- Competence Center for Integrated Software and AI Systems (INTEGRATE) within the Austrian COMET Program and by the German Federal Ministry of Education and Research (BMBF) within the project DeepWrite (Grant. No. 16DHBKI059).
\todo[inline]{Add FrAInderl ack}

%
%
%
\bibliographystyle{splncs04}
\bibliography{main}

\section*{Appendix}

\begin{table}[H]
\renewcommand\thetable{A.1}
	\centering
	\begin{tabular}{l|c|c}
        \toprule
		Parameter Name  				& Default Value& GAT  \\
        \midrule
		GCN Layer Sizes					&\multicolumn{2}{c}{(256, 256, 256)}\\\hline
            Attention Heads                 &-&2\\\hline
		Epochs							&\multicolumn{2}{c}{200}	\\\hline
		Dropout	Probability				&\multicolumn{2}{c}{0.2}\\\hline
		Learning Rate			        &\multicolumn{2}{c}{0.01}	\\
        \bottomrule
	\end{tabular}
	\caption{Default hyper-parameters and model-specific settings.}
	\label{tab:default_parameters}
\end{table}

\end{document}